\documentclass[11pt]{article}

\usepackage[final]{acl}

\usepackage{times}
\usepackage{latexsym}

\usepackage[T1]{fontenc}
\usepackage[utf8]{inputenc}
\usepackage{microtype}
\usepackage{inconsolata}
\usepackage{graphicx}
\usepackage{booktabs}
\usepackage{multirow}
\usepackage{enumitem}
\usepackage{float}
\usepackage{dblfloatfix}
\usepackage{xurl}

\title{Quality over Quantity: Rethinking Data and Model Assumptions for Bengali Coreference Resolution}

\author{Zenith Biswas \\
  Data Science and Artificial Intelligence \\
  Saarland University \\
  Saarbrücken, Germany \\
  \texttt{zebi00001@stud.uni-saarland.de}
  \And
  Andrew Thomas Dyer \\
  Language Science and Technology \\
  Saarland University \\
  Saarbrücken, Germany \\
  \texttt{andrew.dyer@uni-saarland.de}}

\begin{document}
\maketitle

\begin{abstract}
Multilingual transformer models are often assumed to generalise across languages, but their effectiveness for coreference resolution in low-resource settings remains unverified. We investigate this gap in Bengali, a language with over 250 million speakers but limited annotated coreference resources. We evaluate transformer-based pre-trained encoders under multiple settings, including zero-shot, fine-tuning, data augmentation, and cross-lingual transfer. Our results challenge several common assumptions about model generalisation and data efficiency in low-resource NLP. We demonstrate that a small amount of high-quality annotated data significantly outperforms larger mixed-quality datasets (p < 0.01). We find that data augmentation, particularly back-translation, does not improve performance and can degrade results in coreference resolution. Cross-lingual transfer is also inconsistent: transfer from a distant language like English reduces performance, and transfer from the related language Hindi provides no consistent benefit across models. Several widely used multilingual models exhibit collapsed embedding spaces, rendering them unsuitable for similarity-based clustering tasks. These findings highlight the importance of annotation quality over data quantity and provide empirical guidance for developing coreference resolution systems in low-resource languages.
\end{abstract}

\section{Introduction}
\label{sec:introduction}

Coreference resolution, the task of identifying all expressions in a text that refer to the same entity, is fundamental to natural language understanding. End-to-end neural approaches \citep{lee2017end, lee2018higher} and transformer-based pre-trained encoders \citep{joshi2020spanbert} have achieved strong results on the English OntoNotes dataset \citep{pradhan2012conll}. However, these advances depend on large annotated datasets that are unavailable for many low-resource languages. 

Bengali is a clear example of this disparity. Although it has more than 250 million speakers \citep{ethnologue2023}, it remains extremely under-resourced for coreference resolution. Only one gold-standard dataset exists: BenCoref, containing just 122 documents \citep{rohan2023bencoref} covering only four domains. The language also poses specific challenges for coreference, including gender-neutral pronouns and complex honorific systems \citep{sikdar2016generalized, sen2022bangla}.

A common approach to overcoming data scarcity is to leverage multilingual pre-trained transformer-based encoder models, which are assumed to generalise across languages through shared representations \citep{pires2019multilingual, conneau2020unsupervised}. This assumption is supported by strategies such as data augmentation \citep{sennrich2016improving, wei2019eda} and cross-lingual transfer \citep{lauscher2020zero}, which have shown promise for various NLP tasks in low-resource settings \citep{hedderich2021survey}. However, coreference resolution depends on precise span boundaries and fine-grained semantic distinctions. Therefore, the effectiveness of these strategies has yet to be empirically verified for coreference resolution in low-resource settings.

In this paper, we investigate the effectiveness of these strategies for Bengali. We evaluate transformer-based pre-trained encoders under multiple settings, including zero-shot, supervised fine-tuning with varying data compositions, data augmentation via back-translation and masked language model paraphrasing, and cross-lingual transfer from Hindi and English. Our architecture uses a frozen encoder with a trainable MLP classifier, chosen for its stability when the disparity between model capacity and available training data makes full fine-tuning unstable \citep{lee2019freeze, mosbach2021stability}.

Our results challenge several common assumptions about model generalisation and data efficiency in low-resource NLP. We find that training on a small set of gold-standard documents significantly outperforms a larger mixed-quality dataset (p < 0.01), which demonstrates that annotation quality matters more than data quantity. While data augmentation via back-translation may perform well in classification tasks, it degrades performance on gold-standard data for coreference resolution, suggesting that augmentation strategies that alter sentence structure are not well suited for coreference resolution. Cross-lingual transfer yields inconsistent results: English transfer reduces performance despite providing substantially more training data, while Hindi transfer helps only MuRIL-Large, with no consistent benefit across other models. We further find that several widely used multilingual models exhibit collapsed embedding spaces for Bengali, rendering them unsuitable for similarity-based clustering tasks despite their strong performance on standard benchmarks.

This paper makes the following contributions:
\begin{itemize}[noitemsep, topsep=2pt, leftmargin=*]
    \item Empirical evidence that annotation quality outweighs data quantity in low-resource coreference resolution, with a small gold-standard dataset significantly outperforming a larger mixed-quality alternative.
    \item Evidence that data augmentation, particularly back-translation, can degrade performance on coreference resolution.
    \item The finding that cross-lingual transfer yields inconsistent results, with no statistically significant aggregate benefit from either related or distant source languages.
    \item A diagnostic for identifying embedding collapse in pre-trained models, applicable beyond Bengali.
    \item Strong baselines for Bengali coreference resolution across multiple experimental configurations.
\end{itemize}
Our code and experimental configurations are publicly available.\footnote{\url{https://github.com/zenith19/bengali-coref-low-resource}}

\section{Related Work}
\label{sec:related-work}

Coreference resolution has seen significant progress in recent years. Early approaches relied on mention-pair classifiers \citep{soon2001machine}, while more recent work has moved toward end-to-end systems that jointly detect and link mentions \citep{lee2017end, lee2018higher}. The integration of span-focused pre-training via SpanBERT further improved performance \citep{joshi2020spanbert}. However, these systems have been developed and evaluated almost exclusively on large English datasets such as OntoNotes \citep{pradhan2012conll}. These datasets contain thousands of annotated documents, raising the question of whether these approaches remain effective under low-resource constraints.

Most coreference research has focused on English, but recent efforts have begun to address other languages. The CorefUD project integrates coreference annotations into the Universal Dependencies framework across multiple languages \citep{zabokrtsky2022corefud}. \citet{bitew2021lazy} explored translation-based approaches for low-resource coreference by projecting annotations through machine translation. For South Asian languages, \citet{mishra2024multilingual} introduced TransMuCoRes, a silver-standard dataset created by translating English documents and projecting coreference annotations via word alignment. Earlier work by \citet{sikdar2016generalized} proposed a rule-based anaphora resolution framework for Indian languages, but their approach predates the transformer era. For Bengali, the only gold-standard resource is BenCoref \citep{rohan2023bencoref}, a 122-document corpus spanning four domains. Despite the availability of these resources, no systematic evaluation of modern pre-trained models or common low-resource strategies has been conducted for Bengali.

Multilingual transformer models such as mBERT \citep{devlin2019bert}, XLM-RoBERTa \citep{conneau2020unsupervised}, and MuRIL \citep{khanuja2021muril} are widely used for cross-lingual transfer. The assumption is that shared subword vocabularies and joint pre-training produce representations that generalise across languages \citep{pires2019multilingual}. However, \citet{lauscher2020zero} showed significant limitations in zero-shot transfer across typologically diverse languages, and representation quality has been found to vary across languages within the same model \citep{ethayarajh2019contextual}. Language-specific models such as BanglaBERT \citep{bhattacharjee2022banglabert} address this by pre-training on target-language data, but their embedding quality for tasks beyond classification has not been evaluated. In particular, whether these models produce embeddings with sufficient discriminability for similarity-based tasks like coreference clustering remains largely unexplored, particularly in Bengali. 

Data augmentation is a standard strategy for addressing data scarcity. Back-translation, which generates paraphrases by round-tripping through a pivot language, has been shown to be effective for machine translation \citep{sennrich2016improving, edunov2018understanding} and has since been broadly adopted. Simple techniques such as synonym replacement have also improved performance in text classification \citep{wei2019eda}. However, these methods are designed for tasks where span preservation is straightforward. Coreference resolution depends on precise span boundaries and fine-grained semantic relationships. Surface-level text modifications, such as changing sentence structure or replacing words with synonyms, can disrupt the annotations they are meant to preserve. In this paper, we empirically evaluate these assumptions for Bengali coreference resolution under low-resource conditions.

\section{Experimental Setup}
\label{sec:experimental-setup}

We evaluate five transformer encoders across nine training configurations on the BenCoref test set, spanning zero-shot, supervised fine-tuning, data augmentation, and cross-lingual transfer. The following subsections describe the datasets, models, architecture, training configurations, augmentation strategies, and evaluation methodology.

\subsection{Datasets}
\label{sec:datasets}

We use four datasets in our experiments. Table~\ref{tab:datasets} summarises their statistics. 

\textbf{BenCoref} \citep{rohan2023bencoref} is the only publicly available gold-standard coreference dataset for Bengali, containing 122 human-annotated documents across four domains: biography, descriptive, story, and novel. We split these into 41 training, 10 development, and 71 test documents using stratified random sampling to preserve domain proportions. The relatively large test set (58.2\%) ensures robust evaluation, while the small training set reflects the extreme data scarcity that motivates this work. All data is stored in CoNLL-2012 format \citep{pradhan2012conll}.

\textbf{TransMuCoRes Bengali} \citep{mishra2024multilingual} provides 100 silver-standard Bengali coreference documents created by translating English LitBank documents using the NLLB machine translation model and projecting coreference annotations via word alignment. These annotations are inherently noisier than BenCoref due to translation errors and alignment mistakes. We use these documents exclusively for training, ensuring that all evaluation is performed on gold-standard BenCoref data.

\textbf{TransMuCoRes Hindi} \citep{mishra2024multilingual} contains 80 silver-standard Hindi documents created through the same pipeline. We use this data for cross-lingual transfer experiments, as Hindi and Bengali share Indo-Aryan heritage, SOV word order, and pro-drop behaviour.

\textbf{OntoNotes 5.0 English} \citep{pradhan2012conll} is the standard benchmark for English coreference resolution. We randomly subsample 437 documents containing coreference annotations from the training split. English serves as a typologically distant transfer source, contrasting with the linguistically related Hindi. The sampled OntoNotes documents are disjoint from any data used to construct TransMuCoRes, ensuring no overlap between training and evaluation sources.

\begin{table}[t]
\centering
\small
\resizebox{\columnwidth}{!}{
\begin{tabular}{lrrrl}
\toprule
\textbf{Dataset} & \textbf{Docs} & \textbf{Tokens} & \textbf{Mentions} & \textbf{Quality} \\
\midrule
BenCoref (Train) & 41 & 15,234 & 1,847 & Gold \\
BenCoref (Dev) & 10 & 4,012 & 458 & Gold \\
BenCoref (Test) & 71 & 29,364 & 3,183 & Gold \\
TransMuCoRes-BN & 100 & 177,134 & 25,402 & Silver \\
TransMuCoRes-HI & 80 & 167,692 & 20,225 & Silver \\
OntoNotes (subset) & 437 & 324,890 & 38,911 & Gold \\
\bottomrule
\end{tabular}
}
\caption{Dataset statistics. BenCoref is gold-standard (human-annotated); TransMuCoRes datasets are silver-standard (automatically projected).}
\label{tab:datasets}
\end{table}

\subsection{Models}
\label{sec:models}

We conducted a preliminary study evaluating ten pre-trained transformer encoders for Bengali coreference resolution. These models span four categories: multilingual (mBERT, RemBERT, XLM-RoBERTa-Large, XLM-RoBERTa-Base), Indic-focused (MuRIL-Large, MuRIL-Base), Bengali-specific (BanglaBERT-Base, BanglaBERT-Large, Bangla-BERT-Base), and a control (BERT-Base-Uncased, which has no Bengali pre-training data).

Five models were excluded from the main experiments. Three models exhibited collapsed embedding spaces with discriminability gaps below 0.01 between coreferent and non-coreferent mention pairs. These are XLM-RoBERTa-Large, XLM-RoBERTa-Base, and MuRIL-Base. This indicates near-indistinguishable representations, making similarity-based clustering methods unreliable (see Section~\ref{sec:embedding-collapse} for detailed analysis). The remaining two were excluded as redundant: BanglaBERT-Large showed marginal discriminability with lower performance than BanglaBERT-Base despite its larger size, and Bangla-BERT-Base was redundant with BanglaBERT-Base.

The five selected models for the main experiments are: MuRIL-Large \citep{khanuja2021muril}, mBERT \citep{devlin2019bert}, RemBERT \citep{chung2021rethinking}, BanglaBERT-Base \citep{bhattacharjee2022banglabert}, and BERT-Base-Uncased \citep{devlin2019bert}. This selection covers multilingual, Indic-focused, and Bengali-specific pre-training, with BERT-Base-Uncased serving as a control to test whether coreference signal comes from structural patterns in contextual embeddings rather than target-language knowledge.

\subsection{Architecture}
\label{sec:architecture}

All experiments use a frozen encoder with a trainable pairwise classifier. This design choice is motivated by the extreme disparity between model capacity (110M--559M parameters) and available training data (as few as 41 documents). A pilot study comparing frozen, LoRA, and full fine-tuning confirmed that freezing the encoder provides stable performance, while full fine-tuning led to unstable behaviour, including embedding collapse in some runs.

\textit{Span and pair features.} We assume gold-standard mention boundaries, so the task is to link mentions rather than detect them. Each mention is turned into a single vector by taking its first token, its last token, and the average of all its tokens from the frozen encoder, and concatenating the three. This captures both the boundaries of the mention and its overall content. To decide whether two mentions $m_i$ and $m_j$ corefer, we build a pair representation from their two vectors together with their element-wise product and absolute difference; the product highlights features the two share, and the difference measures how far apart they are.

\textit{Classifier.} A three-layer MLP with sigmoid activation produces a coreference score $s_{ij} \in [0, 1]$. The classifier has approximately 4.85M trainable parameters (up to 7.2M for the largest encoders). Only this classifier is trained; the encoders (110M to 559M parameters) are frozen and receive no gradient updates, and the trainable parameters are a small fraction of the full model. This is a deliberate design choice for the low-resource setting. Rather than exposing hundreds of millions of encoder parameters to as few as 41 documents, we keep the pre-trained representations fixed and train only a compact classifier on top, which makes learning feasible at this data scale.

\textit{Clustering.} Coreference clusters are formed by applying connected-components clustering over the pairwise scores. Mention pairs with scores above a threshold are treated as edges in an undirected graph, and the connected components define the predicted clusters. Thresholds are tuned on the development set for each model individually.

\subsection{Training Configurations}
\label{sec:training-configs}
Table~\ref{tab:configs} summarises the training data composition for each experimental configuration. All configurations share the same BenCoref development (10 documents) and test (71 documents) sets, ensuring that performance differences reflect variations in training data only.

\begin{table}[t]
\centering
\small
\resizebox{\columnwidth}{!}{
\begin{tabular}{llr}
\toprule
\textbf{Configuration} & \textbf{Training Data} & \textbf{Docs} \\
\midrule
Zero-Shot & None & 0 \\
Mixed Baseline & BenCoref + TransMuCoRes-BN & 141 \\
Mixed + BT & Mixed + back-translated & 282 \\
Mixed + Para & Mixed + paraphrased & 282 \\
Hindi Transfer & TransMuCoRes-HI & 80 \\
English Transfer & OntoNotes subset & 437 \\
Gold Only & BenCoref only & 41 \\
Gold + BT & BenCoref + back-translated & 82 \\
Gold + Para & BenCoref + paraphrased & 82 \\
\bottomrule
\end{tabular}
}
\caption{Training configurations. BT = back-translation, Para = MLM paraphrasing. All configurations use the same BenCoref dev and test sets.}
\label{tab:configs}
\end{table}

These configurations are designed to isolate the effects of data quality, augmentation, and cross-lingual transfer. Comparing Gold Only against Mixed Baseline tests whether data quality outweighs quantity. Comparing Gold + BT and Gold + Para against Gold Only tests whether augmentation helps on clean data. Hindi and English Transfer test cross-lingual approaches from a related and a distant language respectively.

\subsection{Data Augmentation}
\label{sec:augmentation}

We apply two sentence-level augmentation strategies to the training data. Both keep the CoNLL structure (document and sentence boundaries and the coreference label column), and re-attach the labels to the new tokens by position. As we discuss in Section~\ref{sec:why-augmentation-fails}, when back-translation changes word order or length, these positional labels remain at their original positions while the words move, so a label can land on an unrelated token instead of its original mention.

\textbf{Back-translation} generates paraphrases by translating each Bengali sentence to English and back to Bengali using Google Translate via the \texttt{googletrans} library (v4.0.0rc1).\footnote{\url{https://pypi.org/project/googletrans/}} This process introduces changes in word order, phrasing, and lexical choice, producing globally restructured sentence variants \citep{sennrich2016improving, edunov2018understanding}.

\textbf{MLM paraphrasing} uses BanglaBERT \citep{bhattacharjee2022banglabert} to generate contextual paraphrases. For each sentence, 20\% of tokens are randomly masked and replaced by sampling from the top-10 predictions of BanglaBERT's masked language modelling head. Unlike back-translation, this approach makes targeted lexical substitutions while preserving the original sentence structure.

Both techniques are applied to the 41-document BenCoref training set (producing 82 documents) and to the 141-document mixed dataset (producing 282 documents).

\subsection{Evaluation}
\label{sec:evaluation}

We use CoNLL F1 as the primary evaluation metric, following the standard established by the CoNLL-2012 shared task \citep{pradhan2012conll}. CoNLL F1 averages three complementary metrics: MUC \citep{vilain1995model}, which evaluates coreference links; B\textsuperscript{3} \citep{bagga1998algorithms}, which evaluates individual mentions; and CEAF\textsubscript{e}~\citep{luo2005coreference}, which evaluates entity-level alignment. No single metric captures all aspects of coreference quality, and the average provides a more robust measure than any individual component.

Statistical significance is assessed using the paired bootstrap test \citep{berg2012empirical} with 10,000 resamples over per-document CoNLL F1 scores. When performing multiple comparisons, we apply the Holm--Bonferroni correction \citep{holm1979simple} to control the family-wise error rate. We report corrected $p$-values throughout.

\section{Results}
\label{sec:results}

\subsection{Model Performance Overview}
\label{sec:model-overview}

Table~\ref{tab:main-results} presents CoNLL F1 scores across all five models and nine training configurations. The best overall result is 66.73\% CoNLL F1, achieved by MuRIL-Large trained on the 41-document gold-standard BenCoref dataset. MuRIL-Large consistently outperforms all other models across configurations, likely due to its Indic-focused pre-training on 17 South Asian languages. The remaining four models perform within a narrow range of 61--64\% CoNLL F1, with no single model consistently ranking second.

Pre-trained models achieve surprisingly strong zero-shot performance (average 63.09\% CoNLL F1) without any task-specific training. BERT-Base-Uncased, which has no Bengali in its pre-training data, achieves the highest zero-shot score (63.82\%), suggesting that structural patterns in contextual embeddings provide useful coreference signals independent of target-language knowledge. Fine-tuning on gold-standard data yields moderate improvements, with MuRIL-Large benefiting most (+4.29 percentage points over its zero-shot baseline).

The averaged CoNLL F1 conceals a systematic imbalance across its components. As shown in Table~\ref{tab:per-metric}, MUC F1 is consistently very high (typically 95 to 97), while CEAF\textsubscript{e} F1 is very low (often 30 to 37), with B\textsuperscript{3} falling in between. Because CoNLL F1 is the mean of the three, a strong MUC score alone can sustain a respectable average. This pattern points to over-linking, since MUC is a link-based metric that is largely insensitive to merging distinct entities, whereas CEAF\textsubscript{e} enforces entity-level alignment and penalises it heavily. Our fine-tuned models reflect this, predicting far fewer clusters than the gold standard (for example, 1.77 against 3.96 average clusters per document for MuRIL-Large under the Mixed configuration). A respectable CoNLL F1 therefore does not, on its own, indicate well-formed clusters. We report the averaged scores for comparability with prior work, and rely on relative comparisons between configurations, which are measured identically and are unaffected by this imbalance.

Error analysis of our best model (MuRIL-Large on gold data) shows that this over-linking is not random but follows a consistent pattern. The model rarely fails to find a true coreference link; instead, it creates far too many false ones, producing 55,107 spurious links for every 1,616 it misses. These spurious links are overwhelmingly between noun phrases, and they typically join mentions that lie far apart in the text: 87\% span more than three sentences, at a mean distance of 20 sentences. A single linguistic phenomenon explains much of this behaviour. Bengali marks respect through gender-neutral honorific pronouns such as \textit{tini} (a formal ``he/she''), which are shared across all prominent, respected referents in a passage. When several such entities co-occur, as in a narrative with multiple characters, their mentions become mutually similar in embedding space, and the model merges them into a single cluster despite referring to distinct people. Long-range over-linking is therefore not generic noise but a predictable consequence of resolving coreference by embedding similarity in a language whose pronoun system withholds the very distinctions the task depends on.

\begin{table}[t]
\centering
\small
\begin{tabular}{lrrrr}
\toprule
\textbf{Model} & \textbf{MUC} & \textbf{B\textsuperscript{3}} & \textbf{CEAF\textsubscript{e}} & \textbf{CoNLL} \\
\midrule
MuRIL-Large      & 96.56 & 66.79 & 36.74 & 66.70 \\
mBERT            & 96.69 & 67.11 & 23.19 & 62.33 \\
RemBERT          & 96.20 & 65.67 & 26.51 & 62.80 \\
BanglaBERT-Base  & 96.34 & 65.67 & 27.23 & 63.08 \\
BERT-Base-Unc    & 95.70 & 66.22 & 27.26 & 63.06 \\
\bottomrule
\end{tabular}
\caption{Per-metric F1 (\%) under the Mixed configuration. High MUC
alongside low CEAF\textsubscript{e} across all models indicates
systematic over-linking, which the averaged CoNLL F1 obscures.}
\label{tab:per-metric}
\end{table}

\begin{table*}[t]
\centering
\small
\resizebox{\textwidth}{!}{
\begin{tabular}{lccccccccc}
\toprule
\textbf{Model} & \textbf{Zero-Shot} & \textbf{Mixed} & \textbf{Mixed+BT} & \textbf{Mixed+Para} & \textbf{Hindi} & \textbf{English} & \textbf{Gold} & \textbf{Gold+BT} & \textbf{Gold+Para} \\
\midrule
MuRIL-Large & 62.44 & 66.70 & 65.06 & 65.12 & 63.96 & 61.35 & \textbf{66.73} & 63.87 & 64.72 \\
mBERT & 63.61 & 62.33 & 63.11 & 63.86 & 63.01 & 61.76 & 64.15 & 64.18 & 64.34 \\
RemBERT & 63.20 & 62.80 & 63.09 & 63.52 & 62.83 & 61.32 & 63.88 & 63.23 & 63.81 \\
BanglaBERT-Base & 62.38 & 63.08 & 63.19 & 63.67 & 63.11 & 61.64 & 63.39 & 63.15 & 63.51 \\
BERT-Base-Unc & 63.82 & 63.06 & 63.11 & 63.27 & 62.71 & 62.26 & 63.64 & 63.28 & 63.06 \\
\midrule
Average & 63.09 & 63.59 & 63.51 & 63.89 & 63.12 & 61.67 & 64.36 & 63.54 & 63.89 \\
\bottomrule
\end{tabular}
}
\caption{CoNLL F1 (\%) across all models and training configurations. Best result in bold. BT = back-translation, Para = MLM paraphrasing. All configurations evaluated on the same BenCoref test set (71 documents).}
\label{tab:main-results}
\end{table*}

\subsection{Gold vs.\ Mixed Data Quality}
\label{sec:gold-vs-mixed}

The most striking finding is that training on 41 gold-standard BenCoref documents (Gold Only) consistently outperforms training on 141 mixed-quality documents (Mixed Baseline) that combine gold and silver-standard data. Gold Only achieves an average CoNLL F1 of 64.36\% compared to 63.59\% for Mixed Baseline. All five models show higher performance with gold-only training. While MuRIL-Large improves only marginally (66.73\% vs.\ 66.70\%), it already achieves its best result with gold data alone. More striking, three models (mBERT, RemBERT, and BERT-Base-Uncased) perform worse with mixed data than their zero-shot baselines, indicating that silver-standard data actively degrades their performance.

The aggregate difference is statistically significant ($p = 0.0070$, paired bootstrap with Holm--Bonferroni correction). This result demonstrates that adding 100 silver-standard documents to 41 gold-standard documents does not provide consistent improvements and can actively harm performance. The noise introduced by automatically projected annotations outweighs the benefit of additional training data, confirming that annotation quality matters more than data quantity in this setting.

The gold and silver datasets, however, differ in more than annotation quality. BenCoref is natively written Bengali, covering biography, descriptive, story, and novel domains, whereas TransMuCoRes-BN is machine-translated from English LitBank fiction \citep{bamman2019annotated}, with its annotations projected automatically through word alignment. So the added noise is not the only difference between them; the genre and the original language differ as well. We do not try to separate these factors, because for Bengali there is currently no cleaner silver dataset to turn to. TransMuCoRes is the most practical option available, and our finding is that adding it did not help and often hurt.

\subsection{Data Augmentation Does Not Improve Performance}
\label{sec:augmentation-results}

Data augmentation does not produce statistically significant improvements in any configuration and actively harms performance in several cases. On gold-standard data, back-translation reduces average CoNLL F1 from 64.36\% to 63.54\% ($-$0.82pp), while MLM paraphrasing produces a smaller decline to 63.89\% ($-$0.47pp). MuRIL-Large is most affected, losing 2.86 percentage points with back-translation and 2.01 with paraphrasing.

Back-translation on gold-standard data produces a statistically significant aggregate degradation ($p = 0.0156$, paired bootstrap with Holm--Bonferroni correction). Paraphrasing shows a smaller, non-significant decline ($p = 0.2076$ after correction).

On mixed data, the pattern is similar. Neither back-translation nor paraphrasing produces a statistically significant change after correction (both p > 0.14). Importantly, MuRIL-Large, the best-performing model, loses over 1.5 percentage points with both augmentation methods on mixed data.

These results suggest that sentence-level augmentation strategies introduce noise that disrupts the precise span boundaries and referential relationships required for coreference resolution. Back-translation, which restructures entire sentences, is more harmful than MLM paraphrasing, which makes only local lexical substitutions.

\subsection{Cross-Lingual Transfer}
\label{sec:transfer-results}

Cross-lingual transfer yields inconsistent results that depend on both the source language and the model. English transfer underperforms the mixed baseline by 1.92pp (corpus-level) and shows no significant difference after correction (p = 0.3260). This occurs despite English providing substantially more training data than any other configuration, suggesting that linguistic mismatch outweighs the benefits of scale.

Hindi transfer (80 TransMuCoRes documents) is approximately neutral on average (63.12\% vs.\ 63.09\% zero-shot, +0.03pp). However, the aggregate effect masks model-specific variation. MuRIL-Large is the only model to benefit meaningfully (+1.52pp), likely because its Indic-focused pre-training on 17 South Asian languages creates better-aligned representations between Hindi and Bengali. Among the remaining four models, BanglaBERT-Base shows a modest improvement (+0.73pp), while mBERT, RemBERT, and BERT-Base-Uncased show small declines (between $-$0.37 and $-$1.11pp). Relative to the mixed baseline (the comparison reported in Table~\ref{tab:significance}), Hindi transfer also shows no significant aggregate effect.

Neither Hindi nor English transfer effects reach statistical significance after Holm--Bonferroni correction ($p = 0.2328$ and $p = 0.3260$ respectively), indicating that the observed differences should be interpreted as trends rather than confirmed findings. The high per-document variance in coreference evaluation limits statistical power to detect transfer effects of this magnitude.

\section{Analysis and Discussion}
\label{sec:analysis}

\subsection{Embedding Collapse}
\label{sec:embedding-collapse}

\begin{figure}[t]
\centering
\includegraphics[width=\columnwidth]{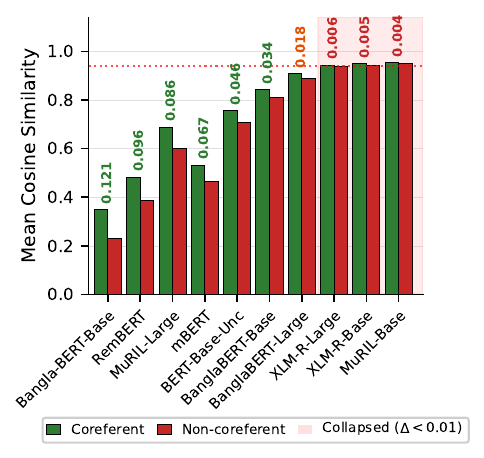}
\caption{Embedding discriminability across all ten candidate models. Green bars show mean within-cluster (coreferent) similarity; red bars show mean between-cluster (non-coreferent) similarity. The gap ($\Delta$) indicates discriminability. Models in the shaded region have gaps below 0.01 and are unsuitable for threshold-based coreference clustering.}
\label{fig:discriminability}
\end{figure}

During the preliminary study, we evaluated embedding quality by measuring the discriminability gap: the difference between mean cosine similarity for coreferent (within-cluster) and non-coreferent (between-cluster) mention pairs. Figure~\ref{fig:discriminability} presents these measurements for all ten candidate models.

Three models exhibit discriminability gaps below 0.01: XLM-RoBERTa-Large, XLM-RoBERTa-Base, and MuRIL-Base. They show mean similarities above 0.94 for both coreferent and non-coreferent pairs. In these models, mention pairs that refer to the same entity are no more similar than pairs that refer to different entities, so their similarity distributions overlap almost completely. No threshold can then group mentions of the same entity together while keeping different entities apart. A low threshold links nearly all pairs and merges every entity into one or two large clusters, whereas a high threshold links almost nothing and leaves most mentions alone in clusters of their own. Neither extreme, nor anything in between, recovers the correct number of clusters (3.96 per document on average in the gold data).

This finding is notable because these three models perform well on standard NLP benchmarks. XLM-RoBERTa is widely used for cross-lingual transfer, and MuRIL-Base shares the same Indic-focused pre-training as MuRIL-Large, which does not exhibit collapse. The divergence between MuRIL-Base and MuRIL-Large suggests that model scale may play a role in preserving embedding discriminability, with larger models maintaining more separable representation spaces.

The discriminability gap is a simple diagnostic that can be computed before any fine-tuning, and we treat it as a necessary condition for similarity-based clustering. A model that fails it would still record a high MUC score, because MUC is largely insensitive to merging distinct entities, and this alone inflates the CoNLL average even though the model has merged separate entities into the same cluster. This is the same over-linking effect we quantify in Section~\ref{sec:model-overview}, in its most extreme form. Running the collapsed models through the full pipeline would only reproduce this artifact at additional cost, so we exclude them on the basis of the diagnostic alone. We recommend that practitioners apply this inexpensive check before using similarity-based methods on a new language, since a model can score well on standard benchmarks yet still map coreferent and non-coreferent pairs to nearly identical vectors.

\subsection{Why Augmentation Fails}
\label{sec:why-augmentation-fails}

Our results show that sentence-level augmentation generally fails to improve performance and often harms it, with back-translation producing a statistically significant degradation. We attribute this to a mismatch between standard augmentation methods and the requirements of coreference resolution.

Back-translation restructures entire sentences by routing text through a pivot language, changing both word order and token count. Our augmentation re-attaches the original coreference labels to the new tokens by position (Section~\ref{sec:augmentation}). As a result, any reordering or change in length shifts labels away from the mentions they were meant to mark. This directly corrupts the span boundaries that coreference resolution depends on. MLM paraphrasing preserves sentence length and structure, so labels stay aligned to their positions; but when a substituted token falls inside a mention, even a plausible synonym can be inappropriate in the mention's context while the label stays unchanged.

More broadly, augmentation methods developed for classification tasks implicitly assume that labels are preserved under surface-level transformations. This assumption holds for sentiment or topic labels, which depend on aggregate semantics, but not for coreference annotations, which depend on specific token spans and their relationships. This suggests that augmentation for coreference, and for other tasks that depend on exact token spans, needs methods that keep labels aligned to their spans, rather than modifying surface text and re-attaching labels by position.

\subsection{Implications for Low-Resource NLP}
\label{sec:implications}

Our findings offer several practical recommendations for researchers working on coreference resolution in low-resource settings. Our evidence comes from a single language and task, so we present the following as recommendations grounded in our setting rather than established generalisations. First, data quality should be prioritised over data quantity. The consistent finding that 41 gold-standard documents outperform 141 mixed-quality documents suggests that investment in careful annotation yields greater returns than collecting larger but noisier datasets. In practice, this means a limited annotation budget is better spent producing a small, carefully checked gold set than a larger set of automatically projected or loosely annotated data.

Second, embedding health should be verified before applying similarity-based methods. Strong performance on standard benchmarks does not guarantee that a model produces usable representations for a target language. The discriminability gap provides a simple diagnostic: it requires only a small set of annotated mention pairs, requires no fine-tuning, and can be computed before committing to a model. For new languages or similar tasks, we recommend this check as a standard step in model selection.

Third, standard augmentation and transfer strategies should not be assumed to generalise. Back-translation and synonym replacement are widely adopted across NLP tasks, but our results show they can actively harm coreference resolution. These methods may behave differently on other span-sensitive tasks such as parsing, so we limit this claim to coreference. Similarly, cross-lingual transfer from a high-resource language like English, often presented as a universal solution, reduced performance in our experiments. Researchers should validate these strategies empirically on their target task rather than relying on results from other settings.

\section{Conclusion}
\label{sec:conclusion}

We presented a systematic study of transformer-based coreference resolution for Bengali under low-resource constraints. Our results challenge three common assumptions in low-resource NLP: that more data is always better, that standard augmentation and cross-lingual transfer reliably help, and that strong benchmark performance implies usable representations for coreference. We show that 41 gold-standard documents outperform 141 mixed-quality documents, that sentence-level augmentation consistently fails to improve performance, and that cross-lingual transfer provides no reliable gains. We identify embedding collapse in widely used multilingual models, demonstrating that benchmark performance does not guarantee discriminative representations for coreference.

These findings highlight that data quality, task-specific validation, and embedding diagnostics matter more than scaling data or model complexity in low-resource settings. Future work could extend these diagnostics to additional languages and explore augmentation methods that preserve span alignment.

\section*{Limitations}
\label{sec:limitations}

Our study has several limitations. We evaluate on a single language. While our findings on data quality, augmentation, and embedding collapse may apply to other low-resource languages, particularly Indo-Aryan ones, we do not verify this empirically.

BenCoref and TransMuCoRes also differ in ways beyond annotation quality: they come from different genres and different source languages. Our comparison therefore reflects the practical choice a Bengali researcher faces today, gold data versus the available silver data. It does not isolate annotation quality as the single cause of the difference, since genre and source language vary as well.

All experiments assume gold-standard mention boundaries, isolating coreference linking from mention detection. End-to-end systems may exhibit different performance patterns. In addition, we use a frozen encoder with a pairwise classifier; alternative architectures or encoder fine-tuning may show different sensitivities to data quality, augmentation, and transfer.

The BenCoref test set (71 documents) is relatively small, which limits statistical power and may explain why several transfer effects did not reach significance.

Finally, we evaluate only two sentence-level augmentation techniques. More advanced methods, such as embedding-level augmentation or approaches that preserve span alignment, may yield different results. Our findings show that standard sentence-level augmentation is ineffective for coreference, not that augmentation is inherently unsuitable for span-sensitive tasks.

\section*{Ethics Statement}

This work uses publicly available datasets (BenCoref, TransMuCoRes, OntoNotes 5.0) in accordance with their respective licenses and intended research use. BenCoref and TransMuCoRes are distributed for research purposes under open licenses; OntoNotes 5.0 is accessed through a valid LDC research license. All pre-trained models evaluated are publicly released on the HuggingFace Hub under permissive licenses. Our contribution is methodological; we do not release any new data. We did not use AI writing assistants for the research itself, though LLM-based tools were used for editorial polishing of the manuscript draft.

\section*{Acknowledgments}
We thank Prof.\ Dr.\ Annemarie Verkerk for her support during this research. We also thank our anonymous reviewers for their helpful suggestions.

\bibliography{custom}

\appendix

\section{Statistical Testing Details}
\label{sec:appendix-stats}

We use the paired bootstrap test \citep{berg2012empirical} with 10,000 resamples to assess statistical significance. For each comparison, per-document CoNLL F1 scores are averaged across all five models to produce 71 independent paired observations (one per test document). The bootstrap procedure resamples these document-level differences with replacement and computes a two-sided p-value as twice the proportion of resamples in which the sign of the mean difference opposes the observed direction.

When multiple comparisons are performed within a family, we apply the Holm--Bonferroni correction \citep{holm1979simple} to control the family-wise error rate. The aggregate tests (Table~\ref{tab:significance}) form one family of 7 comparisons. Per-model tests within each comparison form separate families of 5 comparisons each. Table~\ref{tab:significance} reports differences in per-document-averaged CoNLL F1 (the statistic the paired bootstrap operates on), which differs in magnitude from the corpus-level CoNLL F1 reported in Table~\ref{tab:main-results}. The two measures can diverge when documents vary substantially in mention count.

\begin{table}[!h]
\centering
\small
\resizebox{\columnwidth}{!}{
\begin{tabular}{llrr}
\toprule
\textbf{Comparison} & \textbf{$\Delta$ F1} & \textbf{95\% CI} & \textbf{$p$ (corr.)} \\
\midrule
Gold vs.\ Mixed             & +2.63pp & [+0.94, +4.54] & 0.0070** \\
BT on Gold vs.\ Gold        & $-$0.81pp & [$-$1.41, $-$0.25] & 0.0156* \\
Para on Gold vs.\ Gold      & $-$0.46pp & [$-$0.97, +0.04] & 0.2076 \\
BT on Mixed vs.\ Mixed      & +1.05pp & [+0.09, +2.06] & 0.1470 \\
Para on Mixed vs.\ Mixed    & +0.83pp & [+0.05, +1.65] & 0.1512 \\
Hindi vs.\ Mixed            & +0.99pp & [$-$0.24, +2.28] & 0.2328 \\
English vs.\ Mixed          & +0.92pp & [$-$0.83, +2.85] & 0.3260 \\
\bottomrule
\end{tabular}
}
\caption{Statistical significance of key experimental comparisons. All differences are computed relative to the baseline listed in each row (e.g., ``Hindi vs.\ Mixed'' compares Hindi Transfer to the Mixed Baseline). Paired bootstrap with 10,000 iterations, Holm--Bonferroni corrected. *$p<0.05$, **$p<0.01$.}
\label{tab:significance}
\end{table}

\end{document}